\pdfoutput=1

\documentclass[11pt]{article}

\usepackage{acl}
\usepackage{times}
\usepackage{latexsym}
\usepackage{graphicx}
\usepackage{breqn}

\usepackage[T1]{fontenc}

\usepackage[utf8]{inputenc}

\usepackage{microtype}

\title{Generating Query Focused Summaries without Fine-tuning the Transformer-based Pre-trained Models}

\author{ Deen Abdullah, Shamanth Nayak, Gandharv Suri, \and Yllias Chali \\
University of Lethbridge \\ 
Alberta, Canada}

\begin{document}
\maketitle
\begin{abstract}
Fine-tuning the Natural Language Processing (NLP) models for each new data set requires higher computational time associated with increased carbon footprint and cost. However, fine-tuning helps the pre- trained models adapt to the latest data sets; what if we avoid the fine-tuning steps and attempt to generate summaries using just the pre-trained models to reduce computational time and cost. In this paper, we tried to omit the fine-tuning steps and investigate whether the Marginal Maximum Relevance (MMR)-based approach can help the pre-trained models to obtain query-focused summaries directly from a new data set that was not used to pre-train the models. First, we used topic modelling on Wikipedia Current Events Portal (WCEP) and Debatepedia datasets to generate queries for summarization tasks. Then, using MMR, we ranked the sentences of the documents according to the queries. Next, we passed the ranked sentences to seven transformer-based pre-trained models to perform the summarization tasks. Finally, we used the MMR approach again to select the query relevant sentences from the generated summaries of individual pre-trained models and constructed the final summary. As indicated by the experimental results, our MMR-based approach successfully ranked and selected the most relevant sentences as summaries and showed better performance than the individual pre-trained models.
\end{abstract}

\section{Introduction}

In text summarization, the aim is to summarize a long document considered as the source text and convert it into a summary that holds all the important information from the source text. Some large-scale datasets have been created with document and summary pairs to perform the summarization tasks. However, the majority of these datasets are not prepared for the Query Focused Summarization (QFS) task \cite{abdullah-chali-2020-towards}. Hence, for each new experiment, preparing the large-scale datasets for the QFS task and then training each neural model may result in higher memory requirement and computational cost \cite{xu-lapata-2020-coarse, xu-lapata-2021-generating}. Moreover, fine-tuning a pre-trained model for each large-scale dataset is a vital step that also requires higher computational time and resources. Therefore, we attempt to determine whether we can minimize the computational time and cost by avoiding the steps of fine-tuning the pre-trained models for each new dataset.

The Maximal Marginal Relevance (MMR) is one of the bagging approaches that has been widely used for video and text summarization tasks \cite{lebanoff-etal-2018-adapting, Hoang2012}. Therefore, we planned to use the MMR approach \cite{carbonell1998use} to generate query-focused summaries only from the pre-trained models without performing the task of fine-tuning. At first we used Latent Dirichlet Allocation (LDA) \cite{blei2003latent} which is a topic modeling approach to generate queries for Wikipedia Current Events Portal (WCEP) \cite{ghalandari2020large} and Debatepedia \cite{nema2017diversity} datasets. We performed the QFS task on multiple pre-trained models to get conglomerate summaries for each document using these datasets. Finally, we used the MMR approach to obtain a final summary considering the query and the generated summaries from the previous steps. We hypothesized that the final generated summary is the better summary among all the generated summaries since MMR selects the highly relevant sentences from the generated summaries of all the seven pre-trained models. Therefore, our proposed approach may choose the most query relevant sentences from conglomerate summaries for a given document and combine the information to generate a better summary.

\section{Related Work}
In Natural Language Processing (NLP), researchers have been showing their interest in summarization for several years and implemented different neural architectures to obtain remarkable performances \cite{dong2019unified, see2017get, zaheer2020big}. To perform the different NLP tasks including the summarization, researchers are using different pre-trained language models such as T5 \cite{raffel2020exploring}, XLNet \cite{yang2019xlnet}, GPT-2 \cite{radford2019language}, Pegasus \cite{zhang2020pegasus}, BART \cite{lewis2019bart}, and LED \cite{beltagy2020longformer}. Although some pioneer works focused on query-focused extractive summaries \cite{davis2012occams, feigenblat2017unsupervised, su2021improve, litvak2017query}, unfortunately, the generated summaries from these works could not conquer the problem of maintaining cohesion in the summaries.

The idea of query focused summarization task was first introduced by Dang \cite{dang2005overview}, where the authors showed that a document with a relevant query can be used to generate short summaries \cite{vig2021exploring}. Again, \citet{nema2017diversity} have proposed an attention mechanism to create query focused summaries. Emphasizing the answer relevant to the summary, \citet{su2021improve} have presented the QFS-BART model that generates relevant answer or summary from a given source document by creating query with a question answering model. Some other QFS related researches were performed \cite{abdullah-chali-2020-towards, egonmwan2019cross, aryal2020selection} where \citet{abdullah-chali-2020-towards} focused on generating queries and performed Query focused Abstractive Summarization (QFAS) task by using document sorting and ranking in preprocessing steps. \citet{egonmwan2019cross} proposed a transfer learning approach by using the tasks of Machine Reading Comprehension (MRC) and query focused summarization. \citet{aryal2020selection} focused on the noisy encoder problem and presented the input sequence in a selective approach to generate query focused abstractive summaries.

In this research, we did not consider any training or fine- tuning approaches since our aim was not to achieve any state-of-the-art (SOTA) result on the QFS task. To minimize the high computational time and cost, we attempted to omit the training and fine-tuning steps of pre-trained models. To the best of our knowledge, we are the first to factor out the steps of fine-tuning the pre-trained models for the summarization task. Therefore, we compared the results of the pre-trained models (without fine-tuning) with our proposed MMR-based approach.

\section{Proposed Framework}
Our proposed framework has three sub-modules. First, query generation, then the document ranking and finally, the query focused summarization task.

\subsection{Query Generation}
Inspired by the work of \citet{blei2003latent}, we used the LDA topic modelling approach on the document to extract topics (5 topics) to be considered as the query of that document. We get a mixture of latent topics from a document using this approach. Equation 1 shows the distribution over words of a document for characterizing the latent topics, which we considered as queries.

\begin{equation}
p(\theta, z, w | \alpha, \beta) = p(\theta|\alpha) \prod_{n=1}^{N}p(z_{n}|\theta)p(w_{n}|z_{n}, \beta)
\end{equation}

where $\theta$ denotes the mixture of the latent topics of the document, $\alpha$ and $\beta$ are the joint distribution of $\theta$, $z$ is a set of $N$ topics, and $w$ is a set of $N$ words. We set the hyperparameters as 1 and 5 for the number of topics and words, respectively. A graphical representation of the LDA model is shown in Figure~1.

\begin{figure*}[h]
\centering
\includegraphics{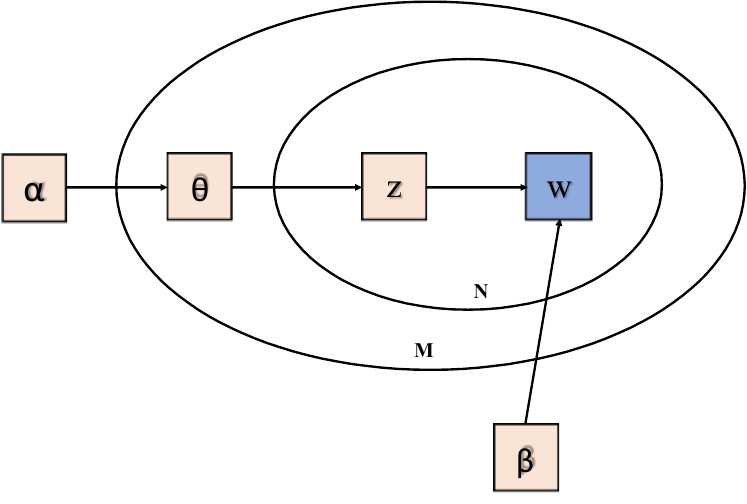}
\caption{Graphical representation of LDA model. Here, M is the number of sentences in a document and N is the number of words in a sentence. The outer circle represents the document, while the inner circle represents the repeated choice of topics and words within a document.}
\end{figure*}

\subsection{Document Ranking}
We used the MMR approach to rank the documents according to the generated query to get the query relevance of the documents. During optimization, the lambda constant for the approach was set to $0.76$. Equation~2 represents the MMR equation for obtaining the query relevant sentences.

\begin{dmath}
MMR = argmax_{D_{i} \in R \backslash S}[\lambda sim_{1} (D_{i}, Q) 
- (1 - \lambda) max_{D_{j} \in S} sim_{2}(D_{i}, D_{j}) ]
\end{dmath}

where, Q is the query, D is the set of documents related to Q, S is the subset of documents in R which are already selected, R $\backslash$ S = set of unselected documents in R and $\lambda$ is the constant in range of $0 - 1$, for diversification of the results.

\subsection{Summarization Task}
After preparing the documents as query-relevant documents, we fed them to seven pre-trained models (LED, T5, Pegasus, BART-AB, XL-Net, BART-EX, GPT2). We produced seven summaries from each document using these seven different models. Then we tokenized all the sentences of the summaries. Finally, we used the MMR approach to select the sentences from these summaries and form the final summary by concatenating the sentences. Sentences were selected based on similarity with the query. The number of sentences to be selected was based on the number of sentences present in the gold summary. The optimal similarity measures were tf-idf for both similarity measures in the MMR algorithm. The lambda constant for the MMR algorithm was 0.83. Overall model architecture is shown in Figure~2.

\begin{figure*}[h]
\centering

\includegraphics{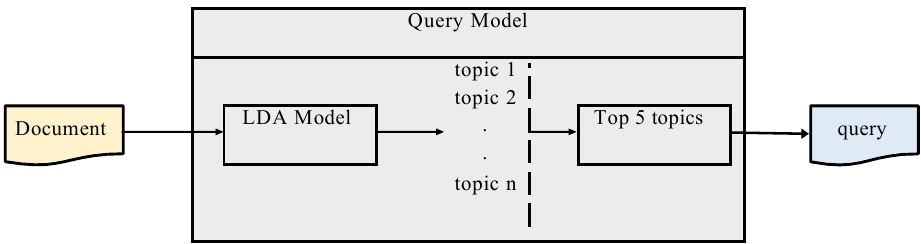}

\vspace*{1cm}

\includegraphics{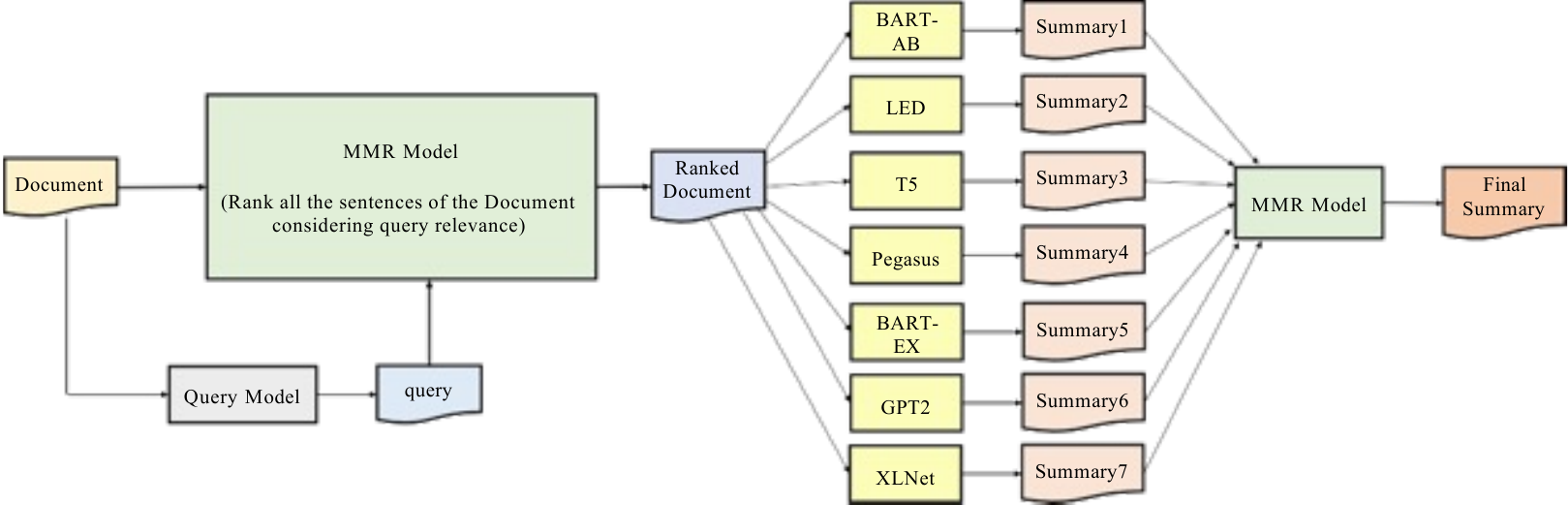}

\caption{Proposed model.}
\end{figure*}

\subsection{Datasets}

\subsubsection{WCEP Dataset}
The Wikipedia Current Events Portal (WCEP) dataset holds contents from Wikipedia, where human-written summaries and corresponding news articles are kept. The authors also stated that they searched for similar articles in the Common Crawl News dataset for each summary and increased the number of source articles. The dataset has a total of 10200 clusters, a total of 2.39M articles where 100 articles per cluster are assigned, and the authors divided the data set for 80\% train, 10\% validation and 10\% for testing purposes \cite{ghalandari2020large}.

\subsubsection{Debatepedia dataset}
The Debatepedia dataset was created to facilitate the query-focused abstractive summarization task. The data set consists of 663 debates under 53 categories. The authors extracted 12695 query, document, summary triples from debatepedia which is an encyclopedia of critical debate topics. The authors distributed 80\% of the data for training, 10\% for validation,  and 10\% for testing purposes \cite{nema2017diversity}.

\section{Experimental Setup}
We used two datasets for our experiment: WCEP (multi-document) and Debatepedia (single document). Since the WCEP dataset consists of 100 documents on each cluster, and many of the documents do not summarize to cover what is present in the gold summary (ground truth summary), we needed to extract those documents which summarize close to the gold summary. We ran an MMR on each cluster with a gold summary as the query for this preprocessing step and selected the top 12 documents from each cluster of the WCEP dataset. We chose MMR for document preprocessing instead of the document-sorting approach on the WCEP dataset, as most of the documents in a cluster contain the same text. While selecting the top 12 documents, a few pairs or triplets (of documents) were present that had the same text. The vectorization of the documents (and the gold summary) was done using pretrained doc2vec. For the similarity measure, we used cosine similarity. The lambda constant used in the MMR algorithm was set at 0.75. Debatepedia dataset is a single document data set and is already preprocessed for the summarization task. 

We used Google Colab (GC)\footnote{https://research.google.com/colaboratory/} as our experimental environment. For the pretrained models, we used Hugging Face \footnote{https://huggingface.co/} on GC. We passed 512 tokens from the ranked document as the input to the models and set the minimum length of the summaries as 15 as the output. We evaluated our proposed approach using ROUGE-1 (R-1), ROUGE-2 (R-2), and ROUGE-L (R-L) \cite{lin2004rouge} to calculate the word-overlap between the reference and the system summaries.

\section{Results}

\begin{table*}[h]
\centering
\begin{tabular}{| c | c | c | c|} \hline
Model & R-1 & 	R-2 & R-L \\ \hline 
BART-AB	& 31.55 & 12.16 & 22.23 \\ \hline
LED & 23.22 & 9.40 & 15.90 \\ \hline
T5 & 28.75 & 9.08 & 20.51 \\ \hline
Pegasus & 32.79 & 12.84 & 23.73 \\ \hline
BART-EX & 16.03 & 7.03 & 11.01 \\ \hline
GPT2 & 16.96 & 7.43 & 11.59 \\ \hline
XLNet & 16.91 & 7.23 & 11.51 \\ \hline
MMR(Our) & 36.89 & 16.15 & 27.94 \\ \hline
\end{tabular}
\caption{ROUGE scores of the models on WCEP dataset}
\end{table*}

Table~1 shows the performance of the pre-trained models on the WCEP dataset. From the table, we can see that the Pegasus model shows the highest performance in R-1, R-2 and R-L scores which is 32.79, 12.84 and 23.73. BART-AB model shows the next highest performance, then T5 and LED, while BART-EX, GPT-2 and XLNet show the lowest performance. After applying the MMR approach, the model successfully outperformed the individual models' performance with R-1, R-2 and R-L scores of 36.89, 16.15 and 27.94, respectively.

\begin{table*}[h]
\centering
\begin{tabular}{| c | c | c | c|} \hline
Model & R-1 & R-2 & R-L  \\ \hline 
BART-AB & 19.47 & 6.67 & 17.93  \\ \hline 
LED & 20.64 & 7.09 & 18.30  \\ \hline 
T5 & 20.08 & 6.47 & 18.09  \\ \hline 
Pegasus & 13.31 & 3.20 & 11.82  \\ \hline 
BART-EX & 19.52 & 6.42 & 16.44  \\ \hline 
GPT2 & 19.75 & 6.63 & 16.61  \\ \hline 
XLNet & 19.27 & 6.17 & 16.08  \\ \hline 
MMR(Our) & 24.79 & 9.05	 & 22.16  \\ \hline 
\end{tabular}
\caption{ROUGE scores of the models on Debatepedia dataset}
\end{table*}

Table~2 shows the performance of the individual pre-trained models on the Debatepedia dataset. From the table, we can see that the LED model shows the highest performance in R-1, R-2 and R-L scores which is 20.64, 7.09 and 18.30. T5, GPT-2, BART-EX, BART-AB, and XLNet models show the next highest performance, while the Pegasus model showed the lowest performance. After applying the MMR approach, the model successfully outperforms the performance of all the individual models with R-1, R-2 and R-L scores of 24.79, 9.05 and 22.16 respectively. 

Based on the results in Table~1 and Table~2, we can say that our MMR-based approach was successful in generating better summaries by combining the summaries of the individual summarization models.

\section{Discussion}

In this research, our goal was to generate queries using the LDA approach for QFS tasks. Next, we ranked all the query- relevant sentences from a source document using the MMR model and performed the summarization task using seven pre-trained models, including BART-AB, LED, T5, Pegasus, BART-EX, GPT-2 and XLNet for WCEP and Debatepedia datasets. Then, we passed all the summaries generated from the pre-trained models to the MMR-based model and finally generated summaries that outperformed the results of the individual models. The MMR model selected the most query relevant sentences as the final summary. We hypothesized that the MMR approach could help to generate better summaries without any training or fine-tuning approach, and with the performance of our approach, we proved our hypothesis. In this work, we did not consider any approach for training or fine-tuning the pre-trained models. Hence, we did not show any comparison of our MMR-based approach with the SOTA models that used training or fine-tuning approaches. However, since our goal was not to achieve a better result than the recent SOTA summarization models, we compared our work with the individual pre-trained models trained with other datasets.

\section{Conclusion}
In this research, one of our aims was to generate queries using the LDA approach and prepare two datasets, such as WCEP and Debatepedia, for the QFS task. Our other goal was to pre-process the source documents and rank the query relevant sentences before passing those sentences to the pre-trained models. Finally, we aimed to use the MMR approach to generate the final summary based on aggregating the summaries of the other seven models. We tried to prove that the MMR-based approach can successfully condense the relevant sentences from different models and generate a better summary for a given document. We found that our MMR-based approach successfully generated a better summary than the summary generated from individual models. Therefore, our proposed MMR-based approach can facilitate the future QFS task and help to obtain better summaries.

 \section{Acknowledgments}
The research reported in this paper was conducted at the University of Lethbridge and supported by the Natural Sciences and Engineering Research Council (NSERC) of Canada discovery grant and the NSERC Alliance - Alberta Innovates Advance Program grant.

\bibliography{anthology,custom}




\end{document}